\documentclass{bmvc2k}

\title{TAG: Boosting Text-VQA via Text-aware\\ Visual Question-answer Generation}

\addauthor{Jun Wang\footnotemark[1]}{junwang@umiacs.umd.edu}{1}
\addauthor{Mingfei Gao}{mingfei.gao@salesforce.com}{2}
\addauthor{Yuqian Hu}{yhu1109@umd.edu}{1}
\addauthor{Ramprasaath R. Selvaraju}{rselvaraju@salesforce.com}{2}
\addauthor{Chetan Ramaiah}{cramaiah@salesforce.com}{2}
\addauthor{Ran Xu}{ran.xu@salesforce.com}{2}
\addauthor{Joseph F. JaJa}{josephj@umd.edu}{1}
\addauthor{Larry S. Davis}{lsd@umiacs.umd.edu}{1}
\addinstitution{
 University of Maryland, \\
 College Park, MD. USA.
}
\addinstitution{
 Salesforce Research, \\
 Palo Alto, CA. USA.
}

\usepackage{bm}
\usepackage{mathtools}
\usepackage{amsmath,amssymb}
\usepackage{dsfont}
\usepackage{siunitx}
\usepackage{adjustbox}
\usepackage{booktabs}
\usepackage{color, colortbl}
\definecolor{Gray}{gray}{0.9}
\usepackage{hyperref}

\ifodd 1
\else

  \newcommand{\com}[1]{}
\fi

\runninghead{Student, Prof, Collaborator}{BMVC Author Guidelines}

\begin{document}

\maketitle
\footnotetext[1]{Part of this work was done when JW was an intern at Salesforce Research.}
\begin{abstract}
Text-VQA aims at answering questions that require understanding the textual cues in an image. Despite the great progress of existing Text-VQA methods, their performance suffers from insufficient human-labeled question-answer (QA) pairs. However, we observe that, in general, the scene text is not fully exploited in the existing datasets-- only a small portion of the text in each image participates in the annotated QA activities. This results in a huge waste of useful information. To address this deficiency, we develop a new method to generate high-quality and diverse QA pairs by explicitly utilizing the existing rich text available in the scene context of each image. Specifically, we propose, TAG, a text-aware visual question-answer generation architecture that learns to produce meaningful, and accurate QA samples using a multimodal transformer. The architecture exploits underexplored scene text information and enhances scene understanding of Text-VQA models by combining the generated QA pairs with the initial training data. Extensive experimental results on two well-known Text-VQA benchmarks (TextVQA and ST-VQA) demonstrate that our proposed TAG effectively enlarges the training data that helps improve the Text-VQA performance without extra labeling effort. Moreover, our model outperforms state-of-the-art approaches that are pre-trained with extra large-scale data. \href{https://github.com/HenryJunW/TAG}{Code is available here}.

\end{abstract}

\section{Introduction}
\label{sec:intro}
Visual question answering (VQA) task \cite{antol2015vqa} aims at inferring the answer to a question based on a holistic understanding of an image. It facilitates many AI applications such as robot interactions \cite{anderson2018vision}, document analysis \cite{mishraICDAR19} and assistance for visually impaired people \cite{bigham2010vizwiz}. 
Text-VQA specifically addresses question answering requests where reasoning text in an image is essential to answer a question. It is a more challenging task in a sense that it requires not only understanding the question and the visual context, but also the embedded text in an image \cite{singh2019towards}.
To achieve this goal, Text-VQA methods \cite{hu2020iterative,kant2020spatially,yang2021tap} aim at studying the interactions among question words, visual objects, and scene text in an image. Recent approaches have focused on either improving transformer-based architectures \cite{vaswani2017attention} in a multi-modal manner \cite{hu2020iterative,kant2020spatially,gao2021structured,liu2020cascade,han2020finding,zhu2021simple,lu2021localize}, or adopting pre-training using additional large-scale data \cite{yang2021tap} to further boost their model performance. All of these methods heavily rely on annotations of question-answer (QA) pairs for model training. Intuitively, the more annotated pairs are leveraged, the better performance a model can achieve. Thanks to the development of text-related VQA datasets \cite{gurari2018vizwiz,wang2021towards,biten2019scene,mishraICDAR19}, Text-VQA has achieved rapid progress.

\begin{figure}[t]
\centering
  \includegraphics[width=0.95\linewidth]{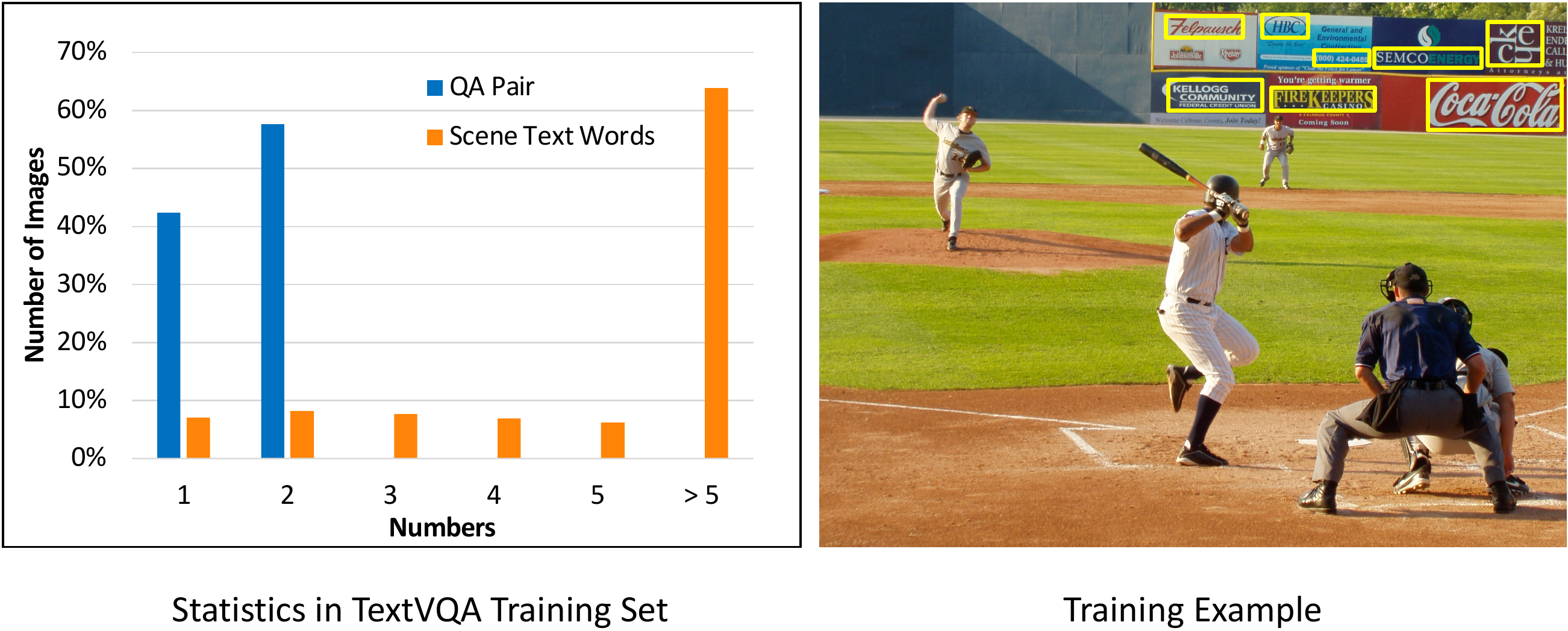}
  \\
  \caption{\textbf{Left:} statistics of the numbers of annotated QA pairs (blue) and scene text words (orange) for each image in the TextVQA training set \cite{singh2019towards}. Clearly, the scene text words are not fully leveraged in the annotations. \textbf{Right:} an example of a training image with more than 5 scene text words, which is typical. Best viewed in color.}
\label{fig:teaser}
\end{figure}

However, the amount of Text-VQA annotations available is still limited due to the sparse labeling of QA pairs in recent datasets. Consider for example the TextVQA dataset \cite{singh2019towards} whose statistics are illustrated in Figure \ref{fig:teaser}. It shows that only one or at most two QA pairs are annotated in the training images. Meanwhile, we also compute the number of text words presented in each image\footnote{We compute the average of different OCR tokens acquired by the Microsoft-OCR system \cite{yang2021tap} for the entire training set.} and observe that most of the images contain at least 5 text words. This observation indicates that scene text is not fully utilized in the annotations, and hence not fully leveraged by recent methods. A natural question would be -- can we fully take advantage of text words in images without incurring extra annotation costs?

As illustrated in Figure \ref{fig:architecture}, we propose to tackle the problem by learning to generate large-scale and diverse text-related QA pairs from existing Text-VQA datasets, using the generated QA pairs to expand the training set and ultimately improving Text-VQA models.
Towards this end, we introduce TAG, a text-aware QA generation model, that generates novel text-related QA pairs at scale. It takes text words (the answer) as one of the inputs and aims at generating a question corresponding to this answer by leveraging the rich visual and scene textual cues. TAG is trained using the originally annotated QA pairs and adapts to generate new QA pairs containing scene text words in images that are not utilized in original annotations. No extra human annotation is required in our framework, so the size and diversity of the training data could be easily and largely increased. Since our generation process is disentangled with the training of Text-VQA models, our generated QA pairs can be used by most of the recent methods.

In summary, we introduce a simple yet efficient text-aware generation approach, which automatically and efficiently generates new QA pairs to improve the performance of the current Text-VQA methods. The main contributions of our work are three-fold: 
\noindent
\begin{itemize}
 \item We identify and analyze possible deficiencies of current Text-VQA datasets-- sparse annotations of QA pairs - and propose to better utilize unused scene text information within each image to improve the model performance.
 
 \item To the best of our knowledge, TAG is the first method that explores scene text-related QA pairs generation for improving Text-VQA tasks without additional labeled data.
 
  \item We consistently demonstrate the effectiveness of our method with two recent Text-VQA models on two Text-VQA datasets. The experimental results suggest that the existing Text-VQA algorithms can benefit from training with the high-quality and diverse QA pairs generated by our method.
\end{itemize}

\section{Related Work}
\label{sec:related_work}
\subsection{Text-related VQA}

To study and evaluate the Text-VQA task, several scene text-based datasets are introduced, including  VizWiz \cite{gurari2018vizwiz}, OCR-VQA\cite{mishraICDAR19}, TextVQA \cite{singh2019towards}, and ST-VQA \cite{biten2019scene}. With the help of these datasets, numerous approaches have been proposed in recent years which increasingly improve Text-VQA performance  \cite{jiang2018pythia,anderson2018bottom,singh2019towards,liu2020cascade,gao2020multi,han2020finding,zhu2021simple,gao2021structured,lu2021localize,hu2020iterative,kant2020spatially,zhang2021position,zeng2021beyond,yang2021tap, Biten_2022_CVPR}. 
LoRRA \cite{singh2019towards} is an early work that extends the original VQA models \cite{jiang2018pythia, anderson2018bottom} with an extra OCR attention branch to select the answer from either a fixed word vocabulary or OCR tokens. Recent studies \cite{devlin2018bert,chen2020uniter,dosovitskiy2020image,carion2020end,liu2021swin,zhao2021point,guan2022m3detr,zhou2020unified,baevski2020wav2vec, wang22n_interspeech,wang2022omnivl} show the benefits of transformer for different vision, language and speech tasks.  M4C \cite{hu2020iterative} develops a transformer-based architecture to fuse different input modalities and iteratively predicts answers through a multi-step answer decoder. Inspired by M4C, more transformer-based models have been proposed with varied structure modifications. Among them, CRN \cite{liu2020cascade} constructs a graph network to model the interactions between text and visual objects. LaAP-Net \cite{han2020finding} predicts a bounding box to explain the generated answer. SSBaseline \cite{zhu2021simple} proposes to split the OCR token features into separate visual and linguistic attention branches. SMA \cite{gao2021structured} reasons over structural text-object graphs and produces answers in a generative way. LOGOS \cite{lu2021localize} introduces a question-visual grounding pre-training task to connect question text and image regions. SA-M4C \cite{kant2020spatially} builds a spatial graph to explicitly model relative spatial relations between visual objects and OCR tokens. TAP \cite{yang2021tap} presents three text-aware pre-training tasks to align representations among scene text, text words, and visual objects. 

However, most of the existing works focus on designing sophisticated architectures that leverage the annotated text in an image and overlook the rich text information that is underused by the annotated QA activities. We fully explore the embedded scene text in images and explicitly generate novel QA pairs that can be used to boost the performance of downstream Text-VQA models.

\subsection{Data Augmentation for VQA}

Data augmentation has been demonstrated to be an effective approach to improve the performance of the VQA task \cite{kafle2017data,shah2019cycle,ray2019sunny,agarwal2020towards,tang2020semantic,wang2021cross,kant2021contrast}. Kafle et al. \cite{kafle2017data} propose to generate new questions using the existing semantic segmentation annotations and templates. Shah et al. \cite{shah2019cycle} introduce a cycle-consistent scheme generating question rephrasings to make VQA models more robust to linguistic variations. Ray et al. \cite{ray2019sunny} propose a consistency-improving data augmentation module to make VQA models answer consistently. Agarwal et al. \cite{agarwal2020towards} explore data augmentation to improve the VQA model's robustness to semantic visual variations. Tang et al. \cite{tang2020semantic} use data augmentation to inject proper inductive biases into the VQA model. Wang et al. \cite{wang2021cross} introduce a generative model for cross-modal data augmentation on VQA. Kant et al. \cite{kant2021contrast} adopt the contrastive loss to make the VQA model robust to linguistic variations in generated questions.  
However, these approaches are designed for the traditional VQA systems that do not emphasize the importance of scene text in their QA tasks. Our method is tailored for the problem of Text-VQA. It takes advantage of the underexploited scene text in images and enlarges the training samples by generating novel text-related QA pairs without the extra labeling cost.

\begin{figure}[t]
\centering
  \includegraphics[width=1.0\linewidth]{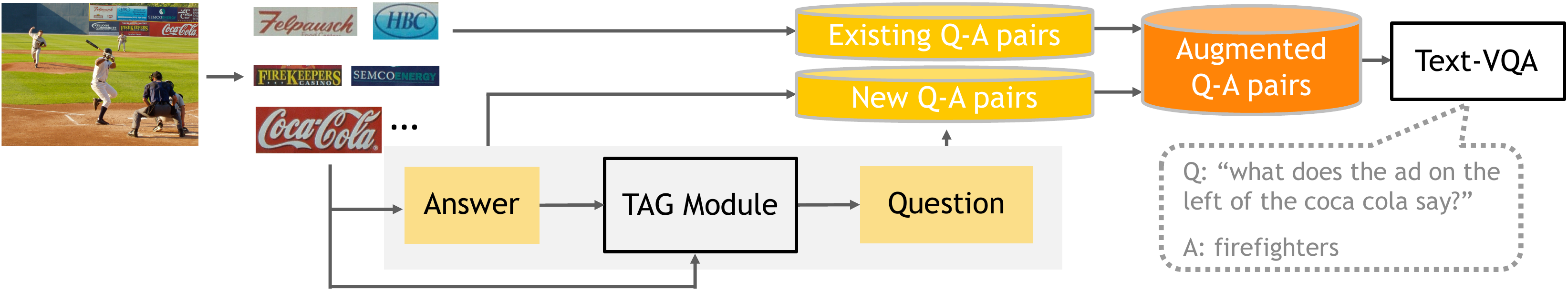}
  \\
  \caption{\textbf{The proposed Text-VQA framework.} It consists of two parts: a text-aware visual question-answer generation module (TAG), followed by a downstream Text-VQA model. TAG is based on a multi-modal transformer architecture, which takes a single image and text words (the answer) as input, and outputs a newly-generated question corresponding to the input answer. The generated QA pairs from TAG together with the originally labeled data are subsequently used to train Text-VQA models, leading to better Text-VQA performance.}
  
\label{fig:architecture}
\end{figure}

\section{Our Approach}
\label{sec:our_approach}
The proposed framework is illustrated in Figure \ref{fig:architecture}, which consists of a transformer-based text-aware visual QA generation module named TAG, followed by a downstream Text-VQA model. Our core module, TAG, carries out text-aware data augmentation tailored for the Text-VQA task and generates novel QA pairs  by leveraging underused scene text in an image. After the TAG module generates a large amount of new QA pairs, we directly augment the training data by combining the generated set and the originally labeled set. The augmented set is used by the downstream Text-VQA models to boost the model performance.

\begin{figure}[t]
\centering
  \includegraphics[width=1.0\linewidth]{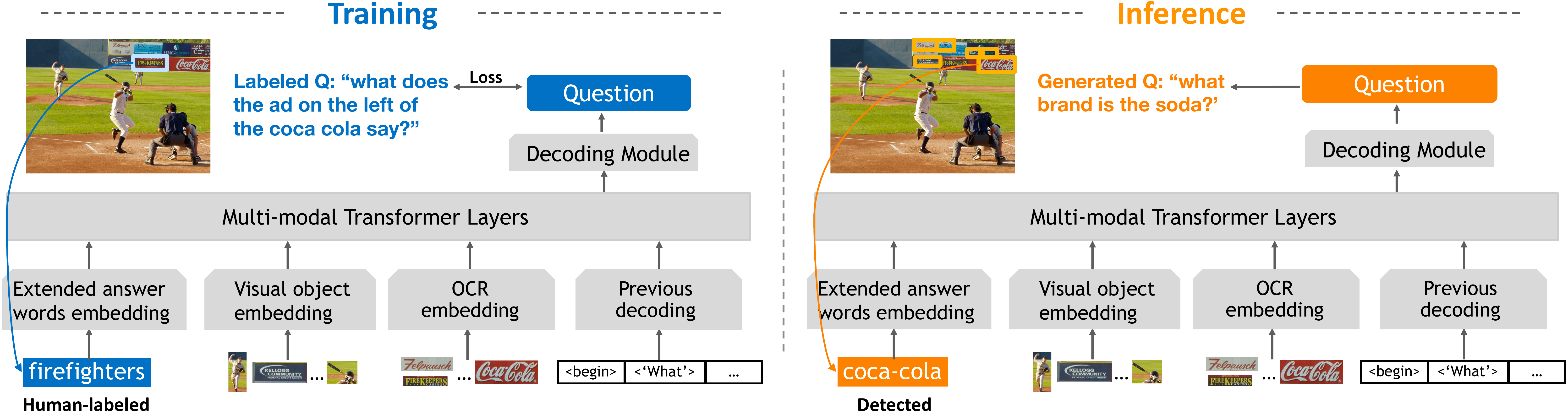}
  \\
  \caption{\textbf{The illustration of our proposed TAG.} High-dimensional feature representations are first extracted for three modalities, including extended answer words, visual objects, and scene text. Then, a multi-modal transformer is used to model the interactions of different modalities. Finally, a decoding module is used to predict the question corresponding to the answer through iterative decoding with an auto-regressive mechanism. \textbf{Left:} Originally labeled QA pairs are used for training. \textbf{Right:} During inference, detected OCR words are used as a novel answer to generate a question. Best viewed in color.}
\label{fig:TAG}
\end{figure}

The workflow of our method is as follows. Given an image, an OCR system and an object detector are used to detect scene text and visual objects, respectively. As illustrated in Figure \ref{fig:TAG}, our TAG takes the scene text words of interest (the answer words), the visual objects and all the detected OCR tokens in the image as inputs and generates a question explicitly corresponding to the answer. Specifically, the answer words, visual objects, and all the OCR tokens are first represented by  high-dimensional features (Section \ref{sec: multi-modal_embedding}).
Then, the multi-modality information is fully aggregated through a transformer architecture with the attention mechanism (Section \ref{sec: multi-modal_fusion}). Finally, the enriched features are used to predict a question to the answer through iterative decoding in an auto-regressive manner (Section \ref{sec: question_prediction}). More details can be found in the supplementary.

\subsection{Multi-modality Feature Embeddings}
\label{sec: multi-modal_embedding}
We describe the feature embedding strategy of our work. The answer words, detected visual objects, and all the detected OCR tokens are embedded as high-dimensional features and then projected into a common d-dimensional embedding space.

\noindent\textbf{Embedding of extended answer words.}
We follow \cite{yang2021tap} to use an extended representation to embed answer words. Given an answer input ${w}^{ans}$, we extend the words with labels of objects ${w}^{obj}$ (detected from the object detector) and scene text OCR words ${w}^{ocr}$ (generated from the OCR system) as a set of $K$ text words.  
A trainable BERT-style model \cite{devlin2018bert} is adopted to extract the embedding of those text words, $\bm{F}^{ans} = \{\bm{f}^{ans}_1,\  \bm{f}^{ans}_2,\ ...,  \bm{f}^{ans}_K \}$, where $k = \{1, 2, ..., K \} $, and $\bm{f}^{ans}_k$ is the d-dimensional feature vectors for $k_{th}$ text word. The embeddings of the set of words are used jointly as the feature of the answer.

\noindent\textbf{Embedding of detected objects}.
Following M4C \cite{hu2020iterative}, we run a pre-trained 2D object detector, Faster R-CNN \cite{ren2015faster} to localize $M$ visual objects for each image. Two visual object features, including appearance and location features are extracted and then combined together to encode each detected object, $\bm{F}^{obj} = \{\bm{f}^{obj}_1,\  \bm{f}^{obj}_2,\ ...,  \bm{f}^{obj}_m \}$, where $m = \{1, 2, ..., M \} $ and $\bm{f}^{obj}_m$ is the projected d-dimensional feature vectors for $m_{th}$ object. Specifically, the feature vector output of the object detector (from the fc7 layer) is used to encode the appearance feature and the relative bounding box coordinates are employed as the location feature. 

\noindent\textbf{Embedding of OCR tokens}. For the $N$ OCR tokens extracted by an OCR system, we construct the embedding for each token containing both its visual and text feature. The visual feature extraction follows the strategy of the above visual object embedding. Additionally, FastText \cite{bojanowski2017enriching} and PHOC features \cite{almazan2014word} are extracted for each OCR token to represent its textual cues. A rich OCR representation is thus obtained, $\bm{F}^{ocr} = \{\bm{f}^{ocr}_1,\  \bm{f}^{ocr}_2,\ ...,  \bm{f}^{ocr}_n \}$, where $n = \{1, 2, ..., N \}$ and $\bm{f}^{ocr}_n$ is the projected d-dimensional feature vectors for $n_{th}$ OCR token.

\subsection{Multi-modality Fusion}
\label{sec: multi-modal_fusion}
Once the feature embedding representation from individual modality, $\bm{F}^{ans}$, $\bm{F}^{obj}$ and $\bm{F}^{ocr}$ are generated, they are able to dynamically attend to each other from a stack of $L$ transformer layers \cite{vaswani2017attention} as shown in Figure \ref{fig:TAG}. The input sequence to the multi-modal transformer is $\bm{F} =\{\bm{F}^{ans}, \bm{F}^{obj}, \bm{F}^{ocr} \}$. The multi-modal transformer leverages feature embeddings from different modalities and accordingly models interaction among them through the multi-head attention mechanism. From the output of the multi-modal transformer, we extract a sequence of d-dimensional feature vectors for each modality, which is an enriched feature from a joint semantic embedding space.

\subsection{Text-aware Visual Question Prediction}
\label{sec: question_prediction}
With the enriched embedding from the multi-modal transformer, the multi-step decoding module predicts a question to the input answer and iteratively generates the question word by word. At each iterative decoding step, we feed in an embedding of previously predicted words, and then the next output word could be either selected from the fixed frequent word vocabulary or from the extracted OCR tokens. Similar to \cite{hu2020iterative,yang2021tap}, two special tokens $<begin>$ and $<end>$ are appended to the word vocabulary, where $<begin>$ is used as the input to the first decoding step and $<end>$ indicates the end of the decoding process. Alternatively, the decoding process ends when the maximum number of steps $T$ is reached.

During training, our TAG is supervised with the binary cross-entropy loss applied using the originally annotated QA pairs and adapts to generate novel QA pairs during generation.
During the QA pairs generation process, we pass an input answer, each of which is selected from the extracted OCR tokens, into the TAG module and generate the corresponding question accordingly. In this way, the generated QA pairs cover a diverse set of scene text which was not directly exploited in the original annotation set. For answer selection, we perform a simple yet efficient strategy that is feeding the OCR token with the largest bounding box as the answer candidate to the proposed TAG. The intuition behind this design is that the scene text with the largest bounding box region is likely to encode semantically meaningful information for scene text-based understanding and reasoning. Also, scene text with a larger font size has a higher chance to be detected correctly without recognition error in general. As we illustrate in our experiments, our simple design facilitates a better understanding of the visual content and provides promising Text-VQA performance. Note that, more high-quality QA pairs could be continuously augmented with a more sophisticated answer-candidate selection strategy. We leave this direction as future work.

\section{Experiments}
We evaluate TAG both qualitatively and quantitatively on the TextVQA \cite{singh2019towards} and the ST-VQA \cite{biten2019scene} datasets. We first present a brief overview of the datasets and implementation details. Then, we empirically validate the effectiveness of our proposed method by comparing it with the existing Text-VQA approaches. Our framework clearly outperforms previous work by a significant margin on both datasets. 

\label{sec:experiments}
\subsection{Datasets and Evaluation Metrics}
\label{dataset}
\noindent \textbf{TextVQA dataset} \cite{singh2019towards} is a widely used benchmark for the Text-VQA task. It consists of 28,408 images sourced from the Open Images dataset \cite{kuznetsova2020open}, with human-annotated questions that require reasoning over text in the images. We follow the standard split on the training, validation and test sets \cite{hu2020iterative,yang2021tap}. For each question, the answer prediction is evaluated based on the soft-voting accuracy of 10 human-annotated answers \cite{goyal2017making, hu2020iterative,yang2021tap}.  

\noindent\textbf{ST-VQA dataset} \cite{biten2019scene} is another popular dataset for the Text-VQA task. It contains 23,038 images from multiple sources including ICDAR 2013 \cite{karatzas2013icdar}, ICDAR 2015 \cite{karatzas2015icdar}, ImageNet \cite{deng2009imagenet}, VizWiz \cite{gurari2018vizwiz}, IIIT STR \cite{mishra2013image}, Visual Genome \cite{krishna2017visual}, and COCO-Text \cite{veit2016coco}. 
The standard evaluation protocol on the ST-VQA dataset consists of accuracy and Average Normalized Levenshtein Similarity (ANLS) \cite{biten2019scene}.

\subsection{Implementation Details}
We use PyTorch to implement our TAG\footnote{Our implementation is built upon the codebase: https://github.com/microsoft/TAP.} that is used to augment the initially labeled data. The augmented dataset is used to improve two recent Text-VQA models, M4C \cite{hu2020iterative} and TAP \cite{yang2021tap}. M4C\textsuperscript{\textdagger} is a variant version \cite{yang2021tap} of M4C, where the detected object labels and scene text tokens are also included in the text encoder, which further improves the performance. 

\begin{table}[t]
    \centering
\begingroup
\setlength{\tabcolsep}{5pt}
\begin{tabular}{l l l c c} 
 \hline
Method & OCR system & Extra Data & Val Acc. & Test Acc. \\ 
 \hline

CRN \cite{liu2020cascade}  & Rosetta-en & $\times$  & 40.39 & 40.96 \\
LaAP-Net \cite{han2020finding}  & Rosetta-en & $\times$  & 40.68 & 40.54 \\
SMA \cite{gao2021structured}  &  SBD-Trans OCR  & $\times$  & 43.74 & 44.29 \\
SSBaseline \cite{zhu2021simple}  & SBD-Trans OCR & $\times$  & 43.95 & 44.72 \\
LOGOS \cite{lu2021localize} & Microsoft-OCR & $\times$  & 50.79 & 50.65 \\

M4C\textsuperscript{\textdagger} \cite{hu2020iterative}  &  Microsoft-OCR & $\times$  & 44.50 & 44.75 \\
\rowcolor{Gray}
M4C\textsuperscript{\textdagger} + TAG  &  Microsoft-OCR & $\times$  & \textbf{45.68} & \textbf{45.96} \\

 TAP \cite{yang2021tap}  & Microsoft-OCR & $\times$  & 49.91 & 49.71 \\
\rowcolor{Gray}
TAP + TAG & Microsoft-OCR & $\times$  & \textbf{52.54} & \textbf{52.57} \\
\hline
LaAP-Net \cite{han2020finding}  & Rosetta-en & ST-VQA  & 41.02 & 41.41 \\
SA-M4C \cite{kant2020spatially}  & Google-OCR & ST-VQA  & 45.40 & 44.60 \\
SMA \cite{gao2021structured}  & SBD-Trans OCR & ST-VQA  & 44.58 & 45.51 \\
SSBaseline \cite{zhu2021simple}  & SBD-Trans OCR & ST-VQA  & 45.53 & 45.66 \\
LOGOS \cite{lu2021localize} & Microsoft-OCR & ST-VQA  & 51.53 & 51.08 \\
M4C\textsuperscript{\textdagger} \cite{hu2020iterative}  &  Microsoft-OCR & ST-VQA  & 45.22 & - \\
\rowcolor{Gray}
M4C\textsuperscript{\textdagger} + TAG  &  Microsoft-OCR & ST-VQA  & \textbf{46.33} & \textbf{46.38} \\

TAP \cite{yang2021tap}  & Microsoft-OCR & ST-VQA  & 50.57 & 50.71 \\
\rowcolor{Gray}
TAP + TAG & Microsoft-OCR & ST-VQA  & \textbf{53.63} & \textbf{53.69} \\

\hline
\end{tabular}
\vspace*{2mm}
\caption{\textbf{TAG's outperformance on the TextVQA dataset when trained on original and augmented dataset under two settings.} Note that M4C\textsuperscript{\textdagger} is the improved version from \cite{yang2021tap}.}
\label{tab:textvqa_results}
\endgroup

\end{table}

\begin{figure}[ht]
\centering
  \includegraphics[width=1\linewidth]{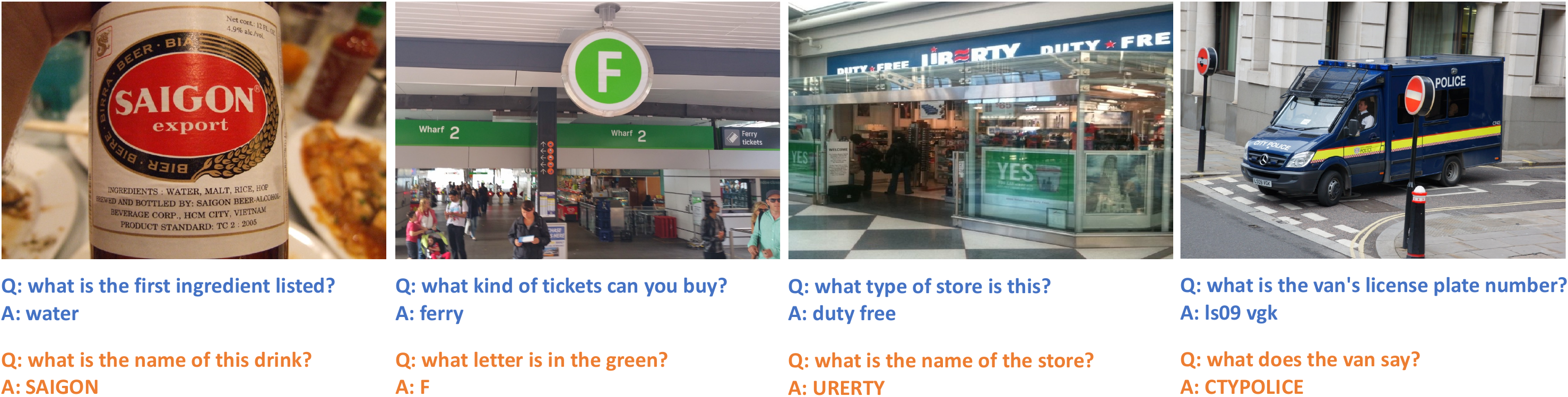}
  \\
  \caption{We visualize the examples of the generated QA pairs (bottom in orange) by TAG module compared with the original annotated QA pairs (top in blue) on the TextVQA training set. "Q" and "A" refer to question and answer, respectively. Best viewed in color.}
\label{fig:visualization_TAG}
\end{figure}

\begin{table}[t]
    \centering
\begingroup
\setlength{\tabcolsep}{5pt}
\begin{tabular}{l l c c c} 
 \hline

Method & Extra Data & Val Acc. & Val ANLS & Test ANLS \\ 
 \hline

CRN \cite{liu2020cascade}   & $\times$ & -  & - & 0.483 \\
LaAP-Net \cite{han2020finding}   & $\times$ & 39.74  & 0.497 & 0.485 \\
SMA \cite{gao2021structured}  & $\times$ & -  & - & 0.486 \\
SA-M4C \cite{kant2020spatially}  &  $\times$ & 42.23  & 0.512 & 0.504 \\
SSBaseline \cite{zhu2021simple}  & $\times$  & - & - & 0.509 \\
LOGOS \cite{lu2021localize} &  $\times$ & 44.10 & 0.535 & 0.522 \\
M4C\textsuperscript{\textdagger} \cite{hu2020iterative}  &  $\times$ & 42.28  & 0.517 & 0.517 \\
\rowcolor{Gray}
M4C\textsuperscript{\textdagger} + TAG  &  $\times$ &  \textbf{44.52}  & \textbf{0.540} & \textbf{0.529}\\

TAP \cite{yang2021tap}   &  $\times$ & 45.29 & 0.551 & 0.543 \\
\rowcolor{Gray}
TAP + TAG   &  $\times$ & \textbf{50.18} & \textbf{0.595} & \textbf{0.586} \\

\hline
SSBaseline \cite{zhu2021simple}  & TextVQA  & - & - & 0.550 \\
LOGOS \cite{lu2021localize} & TextVQA & 48.63 & 0.581 & 0.579 \\

M4C\textsuperscript{\textdagger} \cite{hu2020iterative} &  TextVQA &  46.60  &  0.560  &  0.552 \\
\rowcolor{Gray}
M4C\textsuperscript{\textdagger} + TAG  &  TextVQA & \textbf{48.69} & \textbf{0.579 } & \textbf{0.571} \\
TAP\textsuperscript{\textdagger}\textsuperscript{\textdagger} \cite{yang2021tap}  &  TextVQA, TextCaps, OCR-CC & 50.83 & 0.598 & 0.597 \\
\rowcolor{Gray}
TAP + TAG &  TextVQA &  \textbf{53.53} & \ \textbf{0.620} & \textbf{0.602} \\

\hline
\end{tabular}
\vspace*{2mm}
\caption{\textbf{Our framework outperforms prior work on the ST-VQA dataset.} Note that M4C\textsuperscript{\textdagger} is the improved version from \cite{yang2021tap}. Specifically, our model with TextVQA outperforms the SOTA approach TAP\textsuperscript{\textdagger \textdagger} \cite{yang2021tap} that is pre-trained with extra large-scale data from external TextCaps \cite{sidorov2020textcaps} and OCR-CC \cite{yang2021tap} datasets.}
\label{tab:stvqa_results}
\endgroup

\end{table}

\noindent\textbf{TAG.}
We project the multi-modality feature embedding to be $d$ = 768 channels. We extract the embedding of extended answer words using the same trainable structure as $BERT_{BASE}$ \cite{devlin2018bert}. Specifically, we initialize the weights of the model from the first three layers of $BERT_{BASE}$ and eliminate the separate text transformer. In terms of the object embedding, a Faster R-CNN object detector \cite{ren2015faster} pre-trained on the Visual Genome dataset \cite{krishna2017visual} is adopted to extract $M = 100$ top-scoring objects on each image and represents each object with its appearance and location features. The Microsoft-OCR system \cite{yang2021tap} is used to extract OCR tokens per image with each token represented with its appearance, location, FastText \cite{bojanowski2017enriching} and PHOC features \cite{almazan2014word}. The multi-modality fusion module is a four-layer transformer with 12 attention heads, which has the same hyper-parameters as $BERT_{BASE}$. We use $T = 30$ decoding steps to predict the output question word by word in an auto-regressive manner.

\noindent\textbf{Training parameters.}
Experiments are conducted on 4 Nvidia P6000 GPUs. We train TAG for 24K iterations with a batch size of 128. We adopt the Adam optimizer \cite{kingma2014adam} with a learning rate of 1e-4 and a staircase learning rate schedule, where we multiply the learning rate by 0.1 at 14K and at 19K iterations. We keep the original parameter settings of downstream Text-VQA models except that we increase their maximum iteration in proportion to the increased size of the augmented data to accommodate the enlarged number of training samples.

\subsection{Main Results}
\label{result}
\noindent\textbf{TextVQA dataset.}
To perform a fair comparison with prior work, we conduct experiments in both the constrained setting (top part of Table \ref{tab:textvqa_results}) and the unconstrained setting (bottom part of Table \ref{tab:textvqa_results}) on the TextVQA dataset \cite{singh2019towards}.\footnote{The constrained setting means training without extra data and the unconstrained one indicates otherwise.} The number of our augmented training QA examples for Text-VQA is 69.2K compared with 34.6K for the original one. In the constrained setting (top), our TAG improves the corresponding M4C and TAP baselines  by 1.18\% and 2.63\% on the validation set, respectively. We note that although LOGOS \cite{lu2021localize} in Table \ref{tab:textvqa_results} uses an extra grounding dataset with 1.1 million images for pre-training and yet our method performs better. In the unconstrained setting (bottom), TAG further boosts M4C and TAP baselines by 1.11\% and 3.06\% on the validation set, respectively. On the TextVQA test set, TAG also obtains significant performance gains over  existing methods. This validates the effectiveness of TAG. 

 We also visualize the generated QA pairs of our TAG in Figure \ref{fig:visualization_TAG}. It shows that our TAG generates meaningful QA pairs that are novel compared to the originally annotated ones.

\noindent\textbf{ST-VQA dataset.}
We also compare our approach with the state-of-the-art (SOTA) methods under both the constrained setting and the unconstrained setting on the ST-VQA dataset \cite{biten2019scene}. We compute the accuracy and ANLS score as the evaluation metrics. The number of the newly built training QA examples for the ST-VQA task after augmentation is 46.8K compared with 23.4K for the original one. Table \ref{tab:stvqa_results} suggests that TAG achieves SOTA performance and significantly outperforms the baselines. In particular, TAP \cite{yang2021tap} achieves 50.83\%, and 0.598 in terms of the accuracy and ANLS score on the validation set with additional TextVQA and 1.4 million large-scale pre-training data, while TAG improves these results by a significant 2.70\% and 0.022 with only additional TextVQA data. In addition, we submit the prediction results of test set on the ST-VQA test server. The results show that TAG with TAP achieves the SOTA performance with ANLS score of 0.602 on the test set. Without bells and whistles, our approach greatly outperforms the baselines, M4C \cite{hu2020iterative} and TAP \cite{yang2021tap}.

\subsection{Ablation Studies}
We conduct extensive ablation studies to demonstrate the effectiveness of TAG using TAP \cite{yang2021tap} under the constrained setting on the TextVQA validation set. 

\begin{table}[htbp]
    \centering
\begingroup
\setlength{\tabcolsep}{5pt}
\begin{tabular}{c@{\ \ \ \ \ \ } c@{\ \ \ \ \ \ \ } c@{\ \ \ \ \ \ \ }@{\ \ \ } c}
\hline
Ans. &  Obj. & OCR. & Val Acc.   \\ 
 \hline
  \checkmark &     &     &  48.76   \\ 
\checkmark &  &    \checkmark       &  48.95 \\ 
\checkmark &   \checkmark  &   &   49.13  \\ 
    \checkmark  & \checkmark & \checkmark &  \textbf{52.54} \\  

\hline
\end{tabular}
\vspace*{2mm}
\caption{Ablation study of TAG with TAP \cite{yang2021tap} under constrained setting on TextVQA validation set. "Ans.", "Obj." and "OCR." refer to embedding of answer words, detected objects and OCR tokens, respectively.}
\label{tab:ablation_modality}
\endgroup
\end{table}

\begin{table}[t]
    \centering
\begingroup
\begin{tabular}[c]{c@{\ \ \ \ \ \ } c@{\ \ \ \ \ \ \ } c@{\ \ \ \ \ \ \ }@{\ \ \ } c}
\hline
Answer Selection & Val Acc.   \\ 
 \hline

 \textit{random}   &  49.26   \\   
\textit{largest}   &   52.54  \\ 
 \textit{top three}        &  52.73 \\ 
  \textit{top five}        &  52.19 \\ 
\hline
\end{tabular}
\vspace*{2mm}
\caption{Ablation study of TAG with TAP \cite{yang2021tap} under constrained setting on TextVQA validation set. \textit{Random} means a random OCR token is selected as the answer input to TAG, while \textit{top three} means the top three largest OCR tokens are selected.}
\label{tab:answer_selection_TAG}

\endgroup
\end{table}

\label{ablation}
\noindent\textbf{Contribution of each modality in TAG.}
To understand the contribution of different input modalities to the success of TAG, Table.~\ref{tab:ablation_modality} summarizes the performance of our framework when a certain modality is removed. It suggests that when both the visual objects and OCR tokens modalities are removed, the performance of our TAG decreases by 3.78\%. On the other side, when removing the visual objects modality and OCR tokens modality separately, the performance drops by 3.59\% and 3.41\%, respectively.

\noindent\textbf{Impact of the answer selection strategy.} To better explore the performance of our TAG, and understand how different answer selection strategies would affect the model performance, we design several experiments over the choice of input answer selection strategy. Our method adopts the \emph{largest} OCR word as the answer candidate for TAG. We compare this strategy with other possibilities in Table.~\ref{tab:answer_selection_TAG}. The table shows that, if we use a \emph{random} OCR token as the input answer, the performance drops by 3.28\%. On the other hand, if we increase the number of answer candidates by including the top-3 largest OCR tokens to augment the labeled data by 3$\times$, the performance boosts additional 0.19\% as compared to the \emph{largest} strategy while it introduces 3$\times$ training time. To achieve a better balance between training efficiency and accuracy, we consider the OCR token with the \emph{largest} bounding box as our final setting for the input answer to TAG. As we have mentioned previously, more high-quality QA pairs could be continuously augmented with a more sophisticated answer-candidate selection strategy. We leave this direction for future work.

\section{Conclusion}
\label{sec:conclusion}
We propose a novel architecture TAG, a text-aware visual question-answer (QA) generation method to deal with the sparse annotation of existing Text-VQA datasets. Our approach leverages the rich yet underexplored visual and scene text information and directly enlarges the existing training set by generating high-quality and rich QA pairs without extra labeling cost. Without bells and whistles, experimental results show that our generated QA pairs  boost the performance of recent Text-VQA models by a large margin on both TextVQA and ST-VQA datasets.

\clearpage
\bibliography{egbib}

\clearpage

\appendix

\renewcommand{\thefigure}{A\arabic{figure}}
\setcounter{figure}{0}

\setcounter{table}{0}
\renewcommand{\thetable}{A\arabic{table}}

\section{Additional Details of Our Approach}
\noindent\textbf{Text-aware Visual Question Prediction.}
We leverage the powerful capability of the attention mechanism in transformers \cite{vaswani2017attention} to capture the interactions among the extended answer words, visual objects, and OCR tokens. Our decoding module is based on a dynamic pointer network \cite{vinyals2015pointer}, which allows both copying words via pointing, and generating words from a fixed vocabulary obtained from the training set.\footnote{Our decoding module is implemented following the implementation of the decoding module in TAP \cite{yang2021tap}. We keep their default hyper-parameters except otherwise noted.}

\begin{table}[h]
    \centering
\begingroup

\begin{tabular}[c]{l@{\ \ \ \ \ \ } c@{\ \ \ \ \ \ \ } c@{\ \ \ \ \ \ \ }@{\ \ \ } c}
\hline
Hyper-parameters & Value   \\ 
 \hline

Max length of answer words ${w}^{ans}$ &  20   \\ 
Max length  $M$  of visual objects ${w}^{obj}$ &  100   \\ 
Max length $N$  of scene texts ${w}^{ocr}$  &  100   \\ 
Max length $K$ of extended answer words & 220 \\
max length $T$ of decoding step & 30 \\
optimizer & Adam \\
batch size  & 128 \\
max iterations & 24K \\
base learning rate & 1e-4 \\
learning rate steps  & 14K, 19K \\
learning rate decay  & 0.1 \\

\hline
\end{tabular}
\vspace*{2mm}
\caption{Hyper-parameters of TAG.}
\label{tab:Hyperparameters_supp}

\endgroup
\end{table}

\noindent\textbf{Hyper-parameters.}
Table.~\ref{tab:Hyperparameters_supp} overviews the hyper-parameter settings of TAG. We use the original parameter settings of downstream Text-VQA models except that we increase the maximum number of iteration in proportion to the increased size of the augmented data to accommodate the enlarged number of training samples.



\section{Additional Qualitatively Visualization of TAG}
\begin{figure}[ht]
\centering
  \includegraphics[width=1\linewidth]{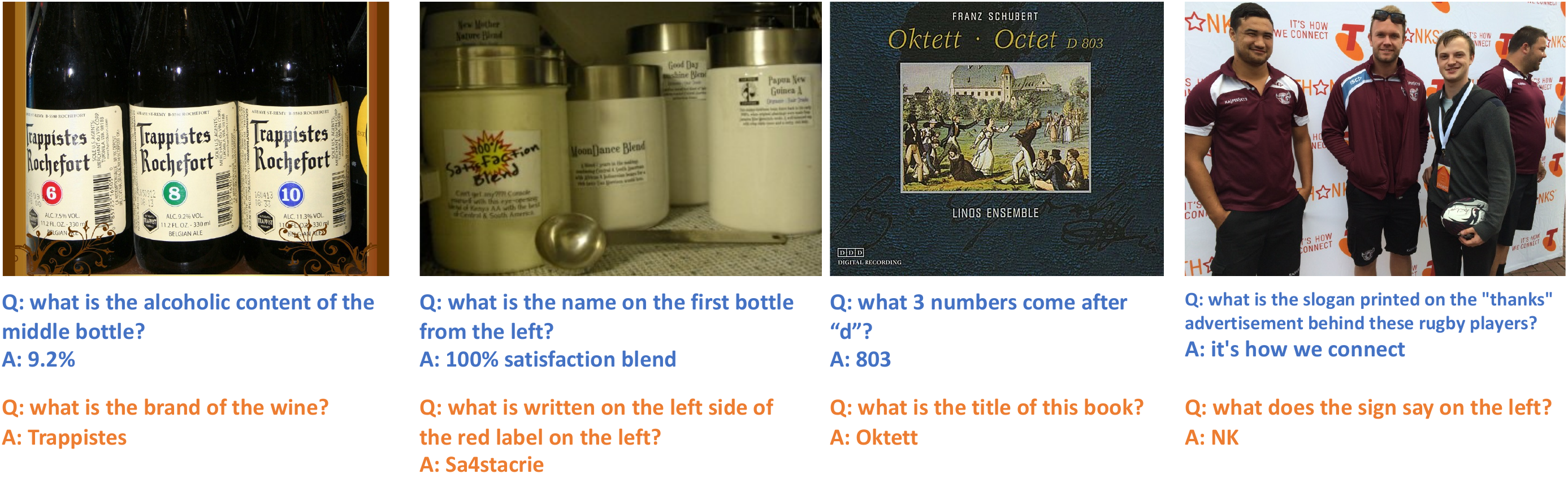}
  \\
  \caption{We visualize the examples of the generated QA pairs (bottom in orange) by the TAG module compared with the original annotated QA pairs (top in blue) on the TextVQA training set. "Q" and "A" refer to question and answer, respectively. Best viewed in color.}
\label{fig:visualization_TAG_supp}
\end{figure}

We present additional visualization results of generated QA pairs by TAG on the TextVQA training set in Figure. \ref{fig:visualization_TAG_supp}. 


\clearpage

\end{document}